\documentclass[conference]{IEEEtran}
\IEEEoverridecommandlockouts
\usepackage{cite}
\usepackage{amsmath,amssymb,amsfonts}
\usepackage{algorithmic}
\usepackage{graphicx}
\usepackage{subcaption}
\usepackage{textcomp}
\usepackage{commath}
\usepackage{xcolor}
\def\BibTeX{{\rm B\kern-.05em{\sc i\kern-.025em b}\kern-.08em
    T\kern-.1667em\lower.7ex\hbox{E}\kern-.125emX}}
\begin{document}

\title{Auto-Encoding Progressive Generative Adversarial Networks For 3D Multi Object Scenes\\
}


\author{\IEEEauthorblockN{Vedant Singh\IEEEauthorrefmark{1},
Manan Oza\IEEEauthorrefmark{2}, Himanshu Vaghela\IEEEauthorrefmark{3} and
Pratik Kanani\IEEEauthorrefmark{4}}
\IEEEauthorblockA{Dwarkadas J. Sanghvi College of Engineering, University of Mumbai\\
Mumbai, India\\
Email: \IEEEauthorrefmark{2} \IEEEauthorrefmark{3}$\{$manan.oza0001, himanshuvaghela1998$\}$@gmail.com,
\IEEEauthorrefmark{1}\IEEEauthorrefmark{4}$\{$vedant.singh, pratik.kanani$\}$@djsce.ac.in}}
\maketitle

\begin{abstract}
3D multi object generative models allow us to synthesize a large range of novel 3D multi object scenes and also identify objects, shapes, layouts and their positions. But multi object scenes are difficult to create because of the dataset being multimodal in nature. The conventional 3D generative adversarial models are not efficient in generating multi object scenes, they usually tend to generate either one object or generate fuzzy results of multiple objects. Auto-encoder models have much scope in feature extraction and representation learning using the unsupervised paradigm in probabilistic spaces. We try to make use of this property in our proposed model. In this paper we propose a novel architecture using 3DConvNets trained with the progressive training paradigm that has been able to generate realistic high resolution 3D scenes of rooms, bedrooms, offices etc. with various pieces of furniture and objects. We make use of the adversarial auto-encoder along with the WGAN-GP loss parameter in our discriminator loss function. Finally this new approach to multi object scene generation has also been able to generate more number of objects per scene.
\end{abstract}


\section{Introduction}
The use of Generative Adversarial Networks (GANs) [1] in creating new images is a active research field nowadays. After 2D images, GANs can also be used in creating 3D models [4, 5, 13, 15, 16, 17] whose applications are increasing and are in great demand. In computer vision, feature detection is an important aspect and there are various available neural networks that do the same. Similarly, in case of 3D objects, recognition of various objects including layouts and shapes is important in various graphics fields.

In creating 3D scenes, it is challenging to create models with fine details because earlier works like Van Kaick et al [2] only considers skeletons of 3D objects into account and looks realistic but fine details are thereby ignored. Due to advancement in deep learning algorithms, and development of large 3-D CAD datasets like Chang et al., 2015 [3], attempts have been made in learning object representations inspired by voxelized objects. New objects are synthesized based on data and it is difficult to do the same due to high dimensionality of 3D shapes and objects as compared to 2D images.

Deep Convolutional Neural Networks are highly efficient in 3D object recognition which can be used fot generative and discriminative purpose in GANs. In [4], single models are generated using 3D GAN and probabilistic spaces. Object structure with significant quality, sampling objects without CAD model and improved 3D object recognition are advantages of this approach over older ones. GANs in 3D object generation provides various advantages like creating objects from probabilistic latent space like uniform distribution or Gaussian distribution, learning distinct features by unsupervised methods instead of learning single feature representation. 

In [5], object classification and shape modelling is feasible by using voxel-based models. A significant improvement in the state-of-art in various applications is observed due to training of voxel based adversarial autoencoder and evaluating models on the ModelNet benchmark which use 2DConvNets that are pre-trained on the ImageNet [20, 21, 22] dataset.

In our approach, we have created a GAN in which layers are increased progressively and is an improvement over [alpha].  Progressive Generative Adversarial Networks [beta] implies that results in 2D images generated can be improved by increasing the dimensions of the data step by step. This concept is further expanded in 3D objects where results obtained are better than previous models.

\section{Related Work}
\subsection{Generative Adversarial Networks}

Generative Adversarial Networks (GAN) used in unsupervised learning is used to generate new images based on input dataset. GAN consists of a generator G and discriminator D. The generative network learns mapping from latent space to a certain data distribution and discriminator differentiates the true data from the output obtained by the generator. The generator’s aim is to increase the error rate of the discriminator. Deep neural network is used to train generator and discriminator. A noise vector $z$ is taken as input by generator which is taken from prior distribution sample $p_z$ and it’s counterfeit sample $G(z)$ is used to fool the discriminator. 
An alternative of GAN named Wasserstein GAN (WGAN) [6] measures Wasserstein distance  between the distribution of the generator and the real distribution. The objective function of WGAN is given as:
\begin{equation}
\min_{G} \max_{D} \mathbb{E}_{x p_d} [D(x)] - \mathbb{E}_{x p_z} [D(G(z))]
\end{equation}
where the distribution of real data is $p_d$.

The newly added GP term by Gulrajani et al. [7] is a critical factor for the discriminator. Therefore, the new GAN objective function is :
\begin{equation}
E_{\hat{x}p_{\hat{x}}} [(\nabla_{\hat{x}} \norm {\hat{x}} - 1)^2]
\end{equation}

where $ p_{\hat{x}}$ is uniform sampling along straight lines among sampled points pairs from $p_d$ and distribution of model $p_g$. It is noted that collapse of mode issue as compared to weight clipping used in original WGAN and  the training is stabilized in WGAN-GP. Therefore, WGAN-GP is used in our proposed model.

\subsection{Progressive Growing of GANs}

GAN network is trained in multiple cycles in progressive GANs [9] .In first cycle, n convolution  layers are added to generate a low resolution output where noise vector $z$  is taken as input. Later, the discriminator is trained with the output generated and the real dataset which is of the same resolution. After the stabilization of training, slightly higher resolution is considered by adding n more convolution layers to the generator and discriminator. This procedure enables model to learn features step by step rather than learning them simultaneously. The output generated with low resolution is more stable and by slowly increasing resolution produces efficient results as compared to simple GAN. 

The training weights are initialised with the weights $N(0,1)$ for each layer at runtime are 
$ \hat{w}_i = w_i/c$ where,
\begin{equation}
c=\bigg(\sqrt{\frac{2}{number of inputs}}\bigg)^{-1}
\end{equation}
The normalized features at every convolutional layers for the generator are given by:
\begin{equation}
b_{x,y}=\frac{a_{x,y}}{\sqrt{\frac{1}{N} \sum\limits_{j=0}^{N-1} (a^j_{x,y})^2 + \epsilon}}
\end{equation}

\subsection{Auto-Encoding GANs for 3D Multi Object Scenes}
Fully Convolutional Refined Auto-Encoding Generative Adversarial Networks for 3D Multi Object Scenes [8] introduces 3D multi objects generation using GANs and SUNCG dataset [10]. An adversarial auto-encoder [18] is combined with GANs in this network. We use an adversarial auto-encoder in place of a variational auto-encoder [19] because adversarial auto-encoders have proven to be generate better results. A refiner is used to refine the generated scenes. Efficient reconstruction performance is obtained due to full convolution and refiner smoothes the scenes giving them mor realistic look.

In our approach progressive approach is applied to this network by increasing resolution of the data step by step hence obtaining better results.

\section{Proposed model}
Our model is a fully convolutional refined auto-encoding progressive generative adversarial network. The network is a combination of a adversarial auto-encoder with generative adversarial networks. The WGAN loss of adversarial auto-encoder is merged with the adversarial auto-encoder using a discriminator as alphaGAN architectures [11]. Additionally the generated scenes are refined by refiner a refiner as done in [12]. We select the shape of the latent space to be 5x3x5x16.
Adversarial auto-encoder allows us to loosen the constraint of distributions and treat this distribution as implicit. Also generator is trained to fool discriminator by generative adversarial network. As a result, this architecture enables reconstruction and generation performance to improve. In addition, refiner allows us to smooth the object shapes and put up shapes to be more realistic visually.

\begin{figure}
  \includegraphics[width=\linewidth]{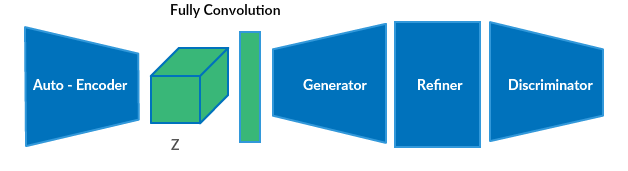}
  \caption{Our proposed architecture. The training samples are passed through an auto encoder and the discriminator that will learn to identify between real ad fake samples. The generator starts from a noise sample $\textit{Z}$ and the generated samples are passed through a refiner. The output of the refiner network is passed through the disciminator to identify whether the given sample is fake or real.}
  \label{fig:boat1}
\end{figure}

\subsection{Input Data}
Contrary to the downscaled dataset used in [8] of size (80$\times$48$\times$80) we use the original SUNCG dataset [10, 14] of dimensions (240$\times$144$\times$240). Here we can see that the later is 27 times bigger than the scenes in the former dataset. We can attribute this as a higher resolution as compared to the dataset used in [8]. Just the way images are made of pixels, voxels are analogous to volumetric displays. It is intuitive that a higher number of voxels in a volumetric display correspond to a higher resolution and thus holds more information. We have eliminated trimming by camera angles and selected scenes that have more than 200,000 voxels. As a result we are left with a dataset of about 190,000 scenes from 10 classes (bedroom, living room, kitchen, room, dining room, office, hall, child room, storage, guest room) with 12 types of objects (empty, ceiling, floor, wall, window, chair, bed, sofa, table, televisions, furniture, miscellaneous objects).

\subsection{Generator}
\begin{figure*}[h!]
\center
  \includegraphics[width=\textwidth]{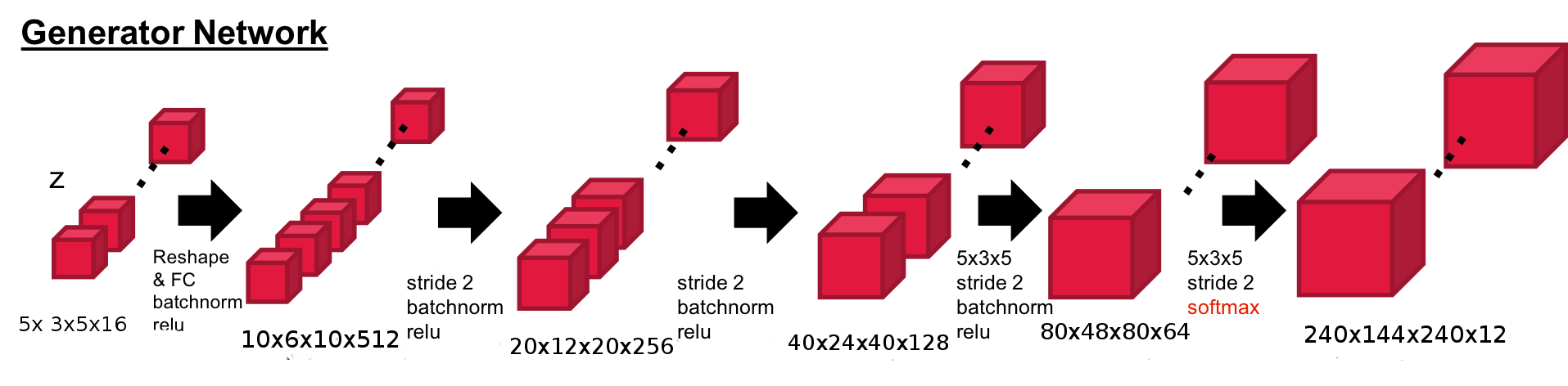}
  \caption{A representation of our generator network. The network starts with a noise vector $  Z$ and is reshaped into a tensor of size (5$\times$3$\times$5$\times$16) and then all the subsequent layers are fully convoluted and the end result is an output tensor having size (240$\times$144$\times$240$\times$12).}
  \label{fig:boat1}
\end{figure*}
The basic architecture of the generator in our proposed model is very much similar to the one used in [8] as shown in Fig. 2. The first layer of latent space is flattened. The alteration is that the last layer has 12 channels with dimensions (240$\times$144$\times$240$\times$12) and is activated by the softmax function. This generator network increments gradually starting with only the first two layersa and after every two epochs the next layer is added. The first layer is fully convolutional and all subsequent layers are 3D convolutional layers with a stride of two voxels and then batch normalization is performed followed by leaky relu activation function.
Fully convolution allows us to extract features more specifically like semantic segmentation tasks. As a result, fully convolution enables reconstruction performance to improve.

\subsection{Encoder}
We use an adversarial auto-encoder in our model.The architecture of the encoder is similar to that of the discriminator network of [8]. The only alteration is that the last layer is 1x1x1 fully convolution.

\subsection{Discriminator}
The discriminator is the exact mirror replica of the generator at each and every step of the training process. The discriminator takes in the outputs generated by generator or the
original dataset as  it's input. It is made up of 3D convolution layers with reshaping done
at the second last layer and the last layer having only one output i.e. true or false.
When the original data is fed into the discriminator, it is trained to identify
between real and fake data and when the output from generator is fed it discriminates
whether the sample provided is real or fake (true or false).

\subsection{Refiner}
The basic architecrure of refiner is similar to SimGAN [12] which is composed with 4 Resnet blocks and 64 channels as shown in Fig. 3.
\begin{figure}[h!]
  \includegraphics[width=\linewidth]{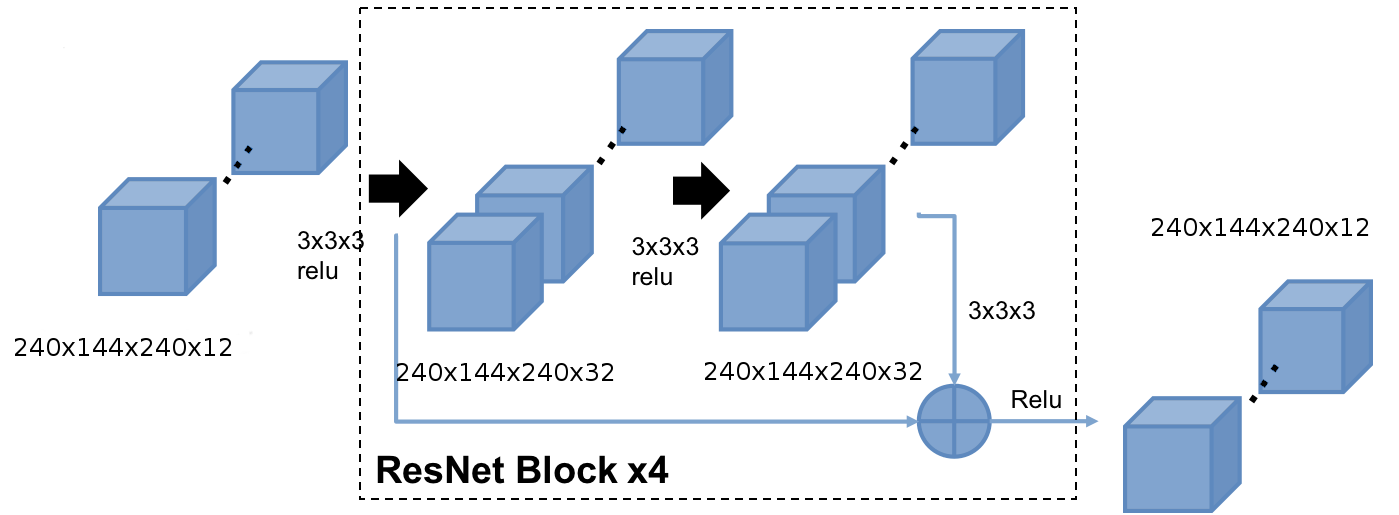}
  \caption{Architecture of the refiner network. Which is same as that used in [8]}
  \label{fig:boat1}
\end{figure}

\section{Analysis}
\subsection{Implementation and training}
We use a cleaned SUNCG dataset with aproximately 190,000 scenes from various categories and rooms.
The reconstruction loss of the network is given as:
\begin{equation}
\begin{aligned}
\textit{L}_{rec} = \sum_{n}^{class} w_{n}(-\gamma x log(x_{rec}) - \\
(1 - \gamma)(1 - x)log(1 - x_{rec}))
\end{aligned}
\end{equation}
$\mathnormal{w}$ represents the occupancy normalized weights with every batch to weight the importance of small objects in the scenes. $\gamma$ is a hyperparameter which weights the relative significance of false positives against false negatives.

The discriminator loss is given as:
\begin{equation}
\begin{aligned}
\textit{L}_{GAN}(D) = -log(D(x)) - log(1 - D(x_{rec})) \\
-log(1 - D(x_{gen})) - E_{\hat{x}p_{\hat{x}}} [(\nabla_{\hat{x}} \norm {\hat{x}} - 1)^2]
\end{aligned}
\end{equation}

The generator loss is given as:
\begin{equation}
\begin{aligned}
\textit{L}_{GAN}(G) = -log(D(x_{rec})) - log(D(x_{gen}))
\end{aligned}
\end{equation}

The optimization takes place as follows:
\begin{itemize}
\item Encoder
\begin{equation}
\begin{aligned}
\min_{E} (\textit{L}_{cGAN}(E) + \lambda L_{rec})
\end{aligned}
\end{equation}

\item Generator and refiner
\begin{equation}
\begin{aligned}
\min_{G} (\lambda \textit{L}_{rec} + L_{GAN}(G))
\end{aligned}
\end{equation}

\item Discriminnator
\begin{equation}
\begin{aligned}
\min_{D} (\textit{L}_{GAN}(D))
\end{aligned}
\end{equation}
\end{itemize}
where $\lambda$ is the hyperparameter that weights the reconstruction loss of the network.

The generator and discriminator are trained progressively meaning that the networks start with 2 layers and these two layers are trained until they reach stabilization and then a new layer is added to both the networks. Our model achieves stability in training each layer for 10 epochs.

\subsection{Results}
We trained our model for aproximately 1,800,000 iterations in total with a batch size of 640 scenes selected at a random out of the entire dataset. The model was trained for aproximately 60 epochs with 3000 iterations per epoch thus summing up to 1,800,000 iterations in total.

Our model was trained progressively starting with only the first two layers and adding each layer after attaining stability. Empirically the model achieved stability for each layer after 10-12 epochs.

One of the samples generated by our model is given below
\begin{figure}[h!]
\centering
  \includegraphics[width=0.5\linewidth]{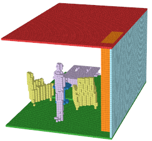}
  \caption{This 3D scene, a sample of generated by our model is of dimensions 240$\times$144$\times$240 voxels.}
  \label{fig:boat1}
\end{figure}
\begin{figure}[h!]
  \centering
  \begin{subfigure}[h!]{1.0\linewidth}
    \includegraphics[width=\linewidth]{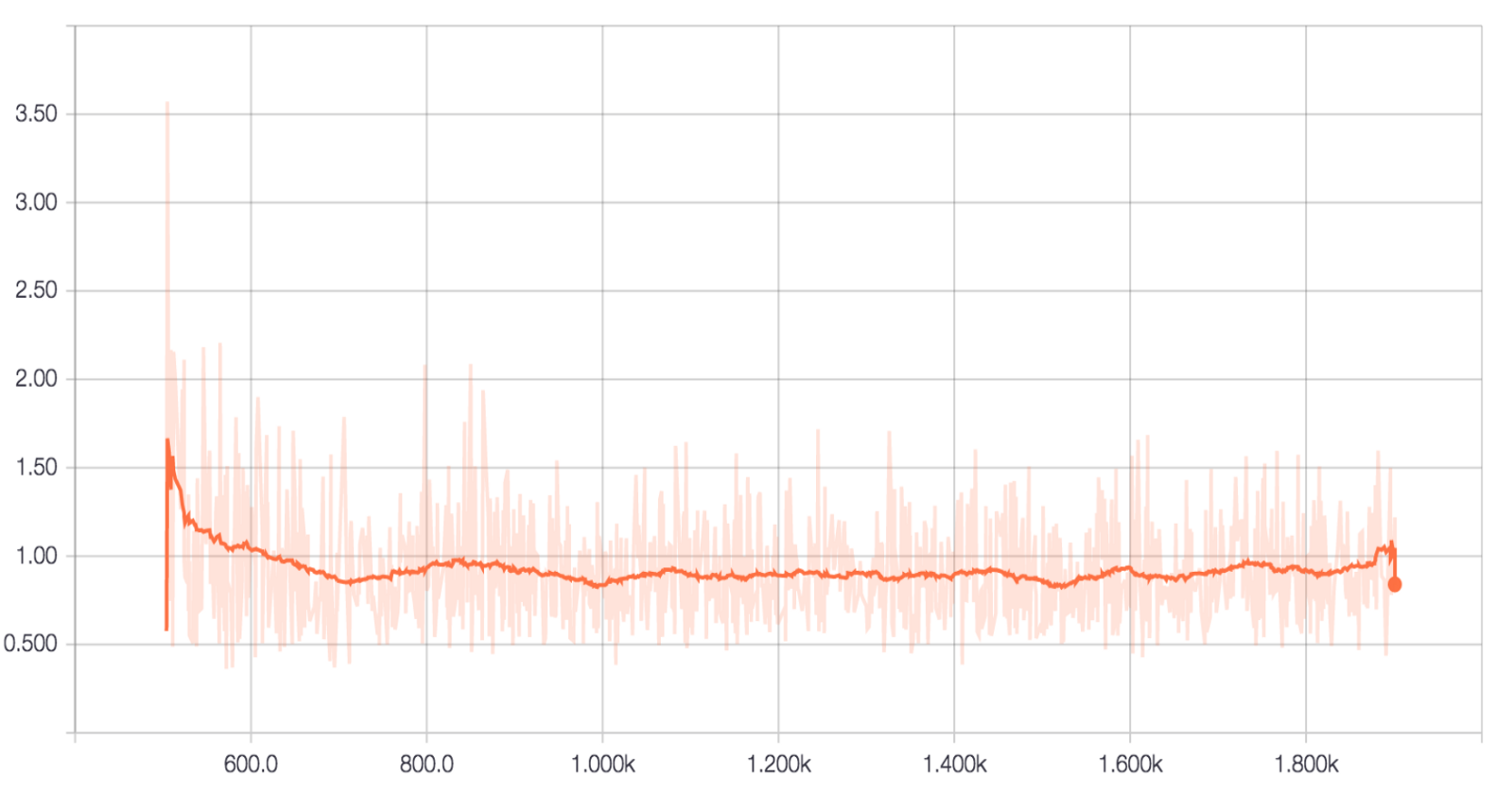}
    \caption{Generator loss.}
  \end{subfigure}
  \begin{subfigure}[b]{1.0\linewidth}
    \includegraphics[width=\linewidth]{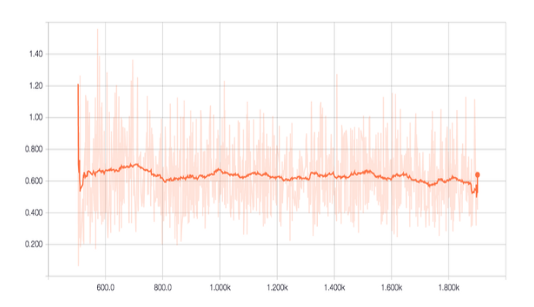}
    \caption{Discriminator loss.}
  \end{subfigure}
  \caption{The plot of generator and discrimiator losses at the end of 1,800,000 iterations (60 epochs).}
  \label{fig:coffee}
\end{figure}
\begin{figure}[h!]
  \centering
  \begin{subfigure}{0.4\linewidth}
    \includegraphics[width=\linewidth]{sample1a.png}
    \caption{}
  \end{subfigure}
  \begin{subfigure}{0.45\linewidth}
    \includegraphics[width=\linewidth]{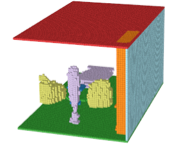}
    \caption{}
  \end{subfigure}  
    \begin{subfigure}{0.45\linewidth}
    \includegraphics[width=\linewidth]{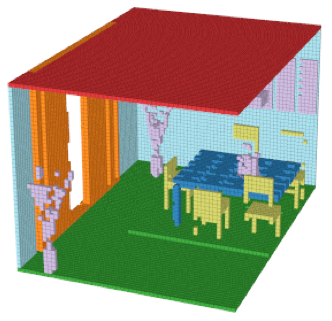}
    \caption{}
  \end{subfigure}
  \begin{subfigure}{0.45\linewidth}
    \includegraphics[width=\linewidth]{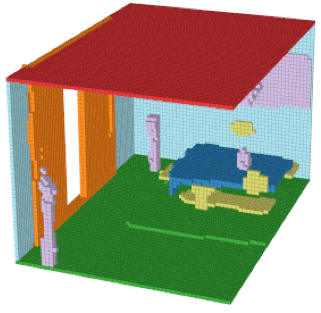}
    \caption{}
  \end{subfigure}
      \begin{subfigure}{0.45\linewidth}
    \includegraphics[width=\linewidth]{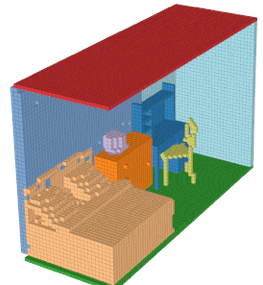}
    \caption{}
  \end{subfigure}
  \begin{subfigure}{0.45\linewidth}
    \includegraphics[width=\linewidth]{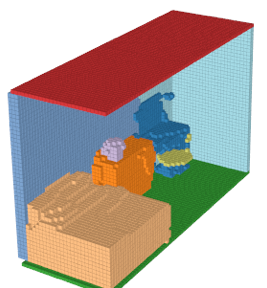}
    \caption{}
  \end{subfigure}
  \caption{The figures on the left are the ones generated by our proposed model while the ones on the right are generated by the model used in [8]. It is evident that our model preserves the object conent inspite of generating higher resolution results.}
\end{figure}

Shown in Fig. 5 are the generator and discriminator losses of our proposed model.

Fig. 6 shows a comparison of the generated samples of our model with the samples of the model given in [8]:

\section{Conclusion}
From the above depicted results our model Auto-Encoding Progressive GAN evidently produces better results than the state of the art 3D-FCR-alphaGAN. Our model is more effective in producing high resolution 3D multi object scenes. Our model also has been successfull in incorporating more number of distinctively identifiable objects in the generated scenes. We have demonstrated that progressive training of the generator and discriminator networks yields sharp and good quality high reolution 3D models. This has further applications in AR/VR and in the animation, computer graphics and gaming industries. The network accuracy is increased but it is equally important to reduce computation time. The future work to this includes increasing novelty in objects, increasing the resolution and make the rendering more realistic and reducing the training time. Microstructures and more intricate details also need to be incorporated by reducing the voxel size.


\begin{thebibliography}{00}
\bibitem{b1} Ian J. Goodfellow et al. Generative adversarial nets. In Proc. NIPS, 2014.
\bibitem{b2} Oliver Van Kaick, Hao Zhang, Ghassan Hamarneh, and Daniel Cohen-Or. A survey on shape correspondence. CGF, 2011.
\bibitem{b3} Angel X Chang, Thomas Funkhouser, Leonidas Guibas, et al. Shapenet: An information-rich 3d model repository. arXiv preprint arXiv:1512.03012, 2015.
\bibitem{b4} Jiajun Wu, Chengkai Zhang, Tianfan Xue, William T. Freeman, Joshua B. Tenenbaum; Learning a Probabilistic Latent Space of Object Shapes via 3D Generative-Adversarial Modeling; arXiv:1610.07584v1.
\bibitem{b5} Andrew Brock, Theodore Lim, J.M. Ritchie, Nick Weston; Generative and Discriminative Voxel Modeling with Convolutional Neural Networks; arXiv:1608.04236v2.
\bibitem{b6} Martin Arjovsky, Soumith Chintala, and Lon Bottou. Wasserstein generative adversarial networks. In Proc. ICML, 2017.
\bibitem{b7} Ishaan Gulrajani, Faruk Ahmed, Martin Arjovsky, Vincent Dumoulin, and Aaron
Courville. Improved training of Wasserstein GANs. In Proc. NIPS, 2017.
\bibitem{b8} Fully Convolutional Refined Auto-Encoding Generative Adversarial Networks for 3D Multi Object Scenes. Yu Nishimura. [online]. https://github.com/yunishi3/3D-FCR-alphaGAN.
\bibitem{b9} Tero Karras, Timo Aila, Samuli Laine, Jaakko Lehtinen, Progressive Growing of
GANs for Improved Quality, Stability, and Variation, In Proc. ICLR, 2018.
\bibitem{b10} Shuran Song, Fisher Yu, Andy Zeng, Angel X. Chang, Manolis Savva, Thomas Funkhouser; Semantic Scene Completion from a Single Depth Image; arXiv:1611.08974v1.
\bibitem{b11} Mihaela Rosca, Balaji Lakshminarayanan, David Warde-Farley, Shakir Mohamed; Variational Approaches for Auto-Encoding Generative Adversarial Networks; arXiv:1706.04987v1.
\bibitem{b12} Ashish Shrivastava, Tomas Pfister, Oncel Tuzel, Josh Susskind, Wenda Wang, Russ Webb; Learning from Simulated and Unsupervised Images through Adversarial Training; arXiv:1612.07828v1.
\bibitem{b13} Christopher B. Choy, Danfei Xu, JunYoung Gwak, Kevin Chen, Silvio Savarese; 3D-R2N2: A Unified Approach for Single and Multi-view 3D Object Reconstruction; arXiv:1604.00449v1.
\bibitem{b14} Princeton University, http://suncg.cs.princeton.edu.
\bibitem{b15} Oliver Van Kaick, Hao Zhang, Ghassan Hamarneh, and Daniel Cohen-Or. A survey on shape correspondence. CGF, 2011.
\bibitem{b16} Johan WH Tangelder and Remco C Veltkamp. A survey of content based 3d shape retrieval methods. Multimedia tools and applications, 39(3):441–471, 2008.
\bibitem{b17} Angel X Chang, Thomas Funkhouser, Leonidas Guibas, et al. Shapenet: An information-rich 3d model repository. arXiv preprint arXiv:1512.03012, 2015.
\bibitem{b18} Alireza Makhzani, Jonathon Shlens, Navdeep Jaitly, Ian Goodfellow, Brendan Frey; arXiv:1511.05644.
\bibitem{b19} Diederik P Kingma, Max Welling; 	arXiv:1312.6114
\bibitem{b20} H. Su, S. Maji, E. Kalogerakis, and E.Learned-Miller. Multi-view convolutional neural networks for 3d shape recognition. In ICCV 2015.
\bibitem{b21} E. Johns, S. Leutenegger, and A. J. Davison. Pairwise decomposition of image sequences for active multi-view recognition. In CVPR 2016.
\bibitem{b22} V. Hegde and R. Zadeh. Fusionnet: 3d object classification using multiple data representations. arXiv Preprint arXiv: 1607.05695, 2016.
\end{thebibliography}
\end{document}